\documentclass[english, letterpaper, 10pt]{article}
\usepackage[T1]{fontenc}
\usepackage{calc}
\usepackage{textcomp}
\usepackage{graphicx}
\usepackage{csquotes}
\usepackage{amssymb,amsmath,amsfonts}
\usepackage{fixltx2e}
\usepackage{fancyhdr}
\usepackage{url}
\usepackage{dsfont}
\usepackage[top=1in, bottom=1.25in, left=0.875in, right=0.875in]{geometry}
\usepackage{bm}
\usepackage{mathptmx}   
\usepackage{titlesec}
\usepackage{titling}
\usepackage{booktabs}
\usepackage{multirow}
\usepackage{subcaption}
\usepackage[justification=centering,font={small}]{caption}
\usepackage[english]{babel}
\usepackage[utf8]{inputenc}
\usepackage{enumitem}
\usepackage[pagebackref=true,breaklinks=true,bookmarks=false,hidelinks]{hyperref}
\usepackage[noabbrev,capitalise]{cleveref}
\usepackage{abstract}
\usepackage{makecell}
\usepackage{icmv}

\usepackage{tikz}
\usetikzlibrary{positioning}

\begin{document}
    \title{How Good MVSNets Are at Depth Fusion}
\author{Oleg Voynov, Aleksandr Safin, Savva Ignatyev, Evgeny Burnaev\\
Skolkovo Institute of Science and Technology,
Bolshoy Boulevard 30, bld.\ 1, Moscow, Russia}
\date{}

\graphicspath{{img/}}
\DeclareGraphicsExtensions{.eps,.pdf,.png,.jpg}

\newcommand{\ra}[1]{\renewcommand{\arraystretch}{#1}}
\newcommand{\cm}[0]{\checkmark}

\ra{1.04}
\setlength{\tabcolsep}{5.5pt}
\def\vs.{vs.\spacefactor=\the\sfcode`\v}
\def\etc.{etc.\spacefactor=\the\sfcode`\c}
\def\wrt.{w.r.t.\spacefactor=\the\sfcode`\c}
\def\ie.{i.e.\spacefactor=\the\sfcode`\c}
\def\eg.{e.g.\spacefactor=\the\sfcode`\c}

\def\parens#1{\left(#1\right)}
\def\braces#1{\left\{#1\right\}}
\def\brackets#1{\left[#1\right]}
\def\norm#1{\left\lVert#1\right\rVert}
\def\abs#1{\left|#1\right|}

    \maketitle

\renewcommand{\abstractname}{\vspace{0pt}\fontsize{11pt}{13pt}\selectfont \uppercase{Abstract}\vspace{-4pt}}
\begin{abstract}
    \normalsize
    We study the effects of the additional input to deep multi-view stereo methods
    in the form of low-quality sensor depth.
    We modify two state-of-the-art deep multi-view stereo methods for using with the input depth.
    We show that the additional input depth may improve the quality of deep multi-view stereo.
    \vspace{6pt}\\
    \textbf{Keywords:} Multi-view stereo, 3D reconstruction, computer vision
\end{abstract}

\section{Introduction}
\label{sec:intro}
3D scanning has a lot of applications.
Autonomous cars or robots need to be able to build a 3d model of the world around them,
to navigate through traffic or an indoor environment.
In computer graphics,
we may want to import the 3d scans of the surroundings into a game or virtual reality,
where we can virtually navigate or interact with the model.

One common way of 3D scan acquisition is depth fusion.
In depth fusion the 3D scan is reconstructed from a set of depth maps,
\ie., images where each pixel represents the distance from the camera to the environment.
The depth maps are captured from different points of view,
for instance, with commodity RGBD cameras such as Microsoft Kinect or Intel RealSense.
Despite impressive real-time reconstruction quality demonstrated by the prior work,
completeness and fine-scale detail remain a fundamental limitation.

Another way of 3D reconstruction is multi-view stereo (MVS).
State-of-the-art MVS methods produce accurate 3D reconstructions from a set of photos
captured from different points of view,
but require distinctive texture for robust matching,
and accurate respective camera parameters, that include positions, orientations and distortions. The methods can be applied within the mobile robot based 3D reconstruction field \cite{newnewnew}.

Although there exist impressive methods combining multi-view photo and depth fusion, such as~\cite{dai2017bundlefusion},
to the best of our knowledge, there is no such method based on the deep learning approach.
At the same time, deep learning has been recently demonstrated to be beneficial for both depth fusion and MVS,
see, \eg.,~\cite{weder2020routedfusion} and \cref{sec:related-mvs}.

In this work, we make a little step towards a deep learning-based combination of MVS and depth fusion,
and study the effects of the additional input to deep MVS methods
in the form of low-quality \textquote{sensor} depth.
To this end,
we modify two state-of-the-art MVS approaches,~\cite{yu2020fast} and~\cite{gu2019cas}, for using with the input depth.

\section{Deep Multi-View Stereo}
\label{sec:related-mvs}
MVS methods are commonly categorized by the output representation,
with the most commonly used representations being (1) voxel grids, (2) point clouds, and (3) depth map sets.
The use of voxel grids is associated with high computational costs and low spatial resolution.
Point clouds are more memory-efficient but are difficult to be processed in parallel
and miss the connectivity information.
The depth maps allow for higher level of detail compared to voxel grids
and possess natural connectivity compared to point clouds.
Depth map representation requires an additional lossy fusion procedure to produce the final reconstruction,
however the current state of the art in MVS is mostly held by the methods using this representation\footnote{
according to \href{https://www.tanksandtemples.org/leaderboard/AdvancedF/}{tanksandtemples.org/leaderboard/AdvancedF}}.

Although among state-of-the-art MVS methods there are still some based on hand-crafted procedures,
like~\cite{xu2019multi},
more and more of them are based on the deep learning approach,
more specifically, convolutional neural networks (CNNs).
The common motivation behind this approach is that
CNNs can use global information for more robust reconstruction in, \eg., low-textured or reflective regions.

One of the first methods based on this approach is MVSNet~\cite{yao2018mvsnet}.
It uses an end-to-end deep architecture,
which first builds a cost volume upon 2D CNN features,
then performs cost volume regularization with a 3D CNN,
and finally refines the output depth maps with a shallow CNN.
Even though this approach shows significant improvement of reconstruction quality
compared to traditional hand-crafted metric based methods,
the memory and computational consumptions grow cubically.

To resolve this issue, the authors of R-MVSNet~\cite{yao2019recurrent}
use convolutional gated recurrent unit for cost volume regularization instead of 3D CNN,
which allows them to reduce the growth of memory consumption from cubic to quadratic.
In parallel, the authors of Point-MVSNet~\cite{chen2019point} propose a coarse-to-fine strategy
to achieve computational efficiency.
They predict a low-resolution 3D cost volume to obtain a coarse depth map
and iteratively upsample and refine it.
Fast-MVSNet~\cite{yu2020fast} further improve the coarse-to-fine strategy
via refinement of the initial coarse cost volume with a differentiable Gauss-Newton layer.

Cas-MVSNet~\cite{gu2019cas} implement the coarse-to-fine strategy in a different way,
via a multi-scale feature extraction followed by several stages of the cost volume calculation.
The key feature of this method is adjustment of the cost volume sampling range
based on semantic information from the features obtained on the current stage.

In this work we investigate the effects of the additional input depth on the quality of the last two methods.
We describe them in more detail in \cref{sec:methods}.

\section{Training Deep MVS}
The most common dataset used for training MVSNets is DTU~\cite{jensen2014large},
containing 124 tabletop scenes.
The data for each scene consists of 49-64 photos captured from different views,
accurate intrinsic and extrinsic camera parameters,
and ground truth point clouds collected with a structured light scanner.

To retrain MVSNets with additional input depth we would ideally need a dataset which is
(1) large-scale, (2) contains low-quality sensor depth and (3) contains accurate ground truth.
To the best of our knowledge, there is no dataset with all three attributes,
so here we list several possible proxies that possess only two.

InteriorNet~\cite{li2018interiornet} contains synthetic RGBD video sequences for 10k indoor environments,
and 15M frames in total.
The low quality sensor depth can be synthesized using a noise model estimated on a real RGBD camera,
\eg., as in~\cite{choi2015robust}.
This dataset is large-scale and contains accurate ground truth but no sensor depth.

ScanNet~\cite{dai2017scannet} contains RGBD sequences for 707 indoor scenes, and 2.5M frames in total.
The depth is captured with Structure sensor,
the camera parameters and ground truth reconstructions are calculated from the sensor depth using
BundleFusion~\cite{dai2017bundlefusion}.
This dataset is large-scale and contains sensor depth but no accurate ground truth.

CoRBS~\cite{wasenmuller2016corbs} contains RGBD video sequences for 4 tabletop scenes, and around 40k frames in total.
The data for each scene consists of 5 sequences captured with Microsoft Kinect v2,
accurate camera parameters from motion caption system,
and ground truth reconstructions collected with a structured light scanner.
This dataset contains both sensor depth and accurate ground truth but only for four simple scenes.

In this work, we used a simple corrupted version of DTU, as we describe in \cref{sec:exp_data}.

\section{Methods}
\label{sec:methods}
The problem of multi-view stereo is,
given a set of images of a scene captured from different points of view
and the respective intrinsic and extrinsic camera parameters,
to reconstruct the 3D model of the scene.
In both methods that we experimented with, FastMVSNet and CasMVSNet,
this problem is split into a set of independent problems
of depth map estimation for one reference image.
The estimated depth maps are then fused into a point cloud with the method of~\cite{galliani2015massively}.

\def\archfigwidth{.8\linewidth}

\subsection{MVSNet}
\label{sec:meth_mvsnet}
\begin{figure*}[t!]
    \centerline{\includegraphics[scale=0.35]{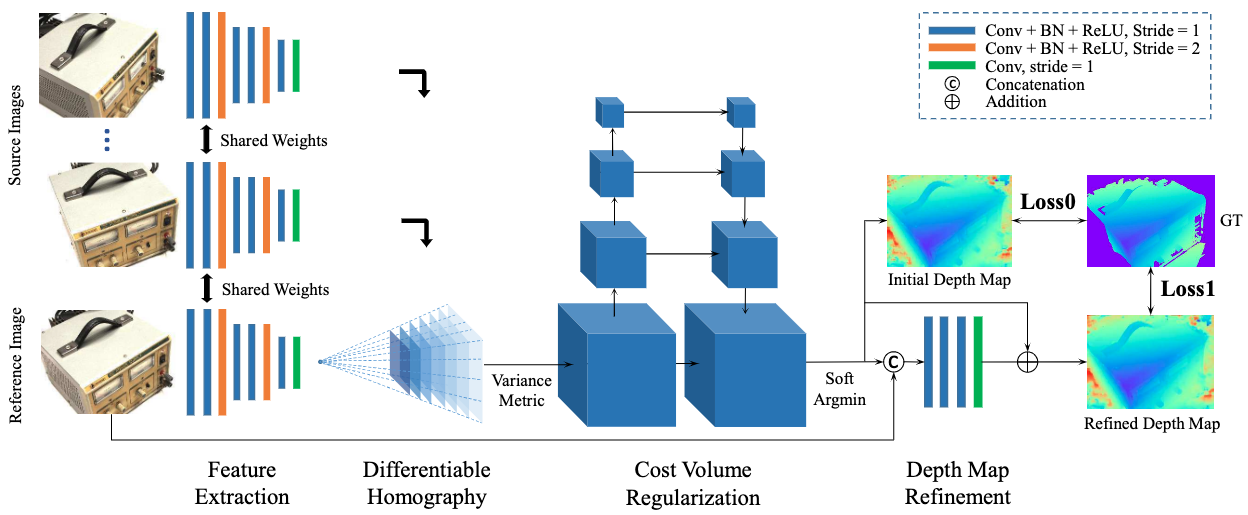}}
    \caption{The architecture of MVSNet.
    A cost volume is constructed upon 2D CNN features and is further regularized with a 3D CNN
    to produce the initial dept map, which is further refined with a shallow CNN.}
    \label{fig:mvs_arch}
\end{figure*}

Since the original MVSNet is the common ground in both methods, we start with its brief description.
MVSNet solves the problem of depth map estimation in the following way (\cref{fig:mvs_arch}).
First, a 2D CNN feature extractor is used to compute features for a reference image \(I_0\)
and several its neighbours \(\braces{I_i}_{i=1}^{N}\).
Then, the extracted features are reprojected to the reference image plane and stacked over a quasi-spatial dimension,
to form the cost volume.
After that, the cost volume is forwarded through a 3D CNN yielding an initial depth map estimation \(d\).
Finally, the initial depth map is refined with a shallow CNN.

\subsection{FastMVSNet}
\label{sec:meth_fast}
\begin{figure*}[t!]
    \centerline{\includegraphics[scale=0.4]{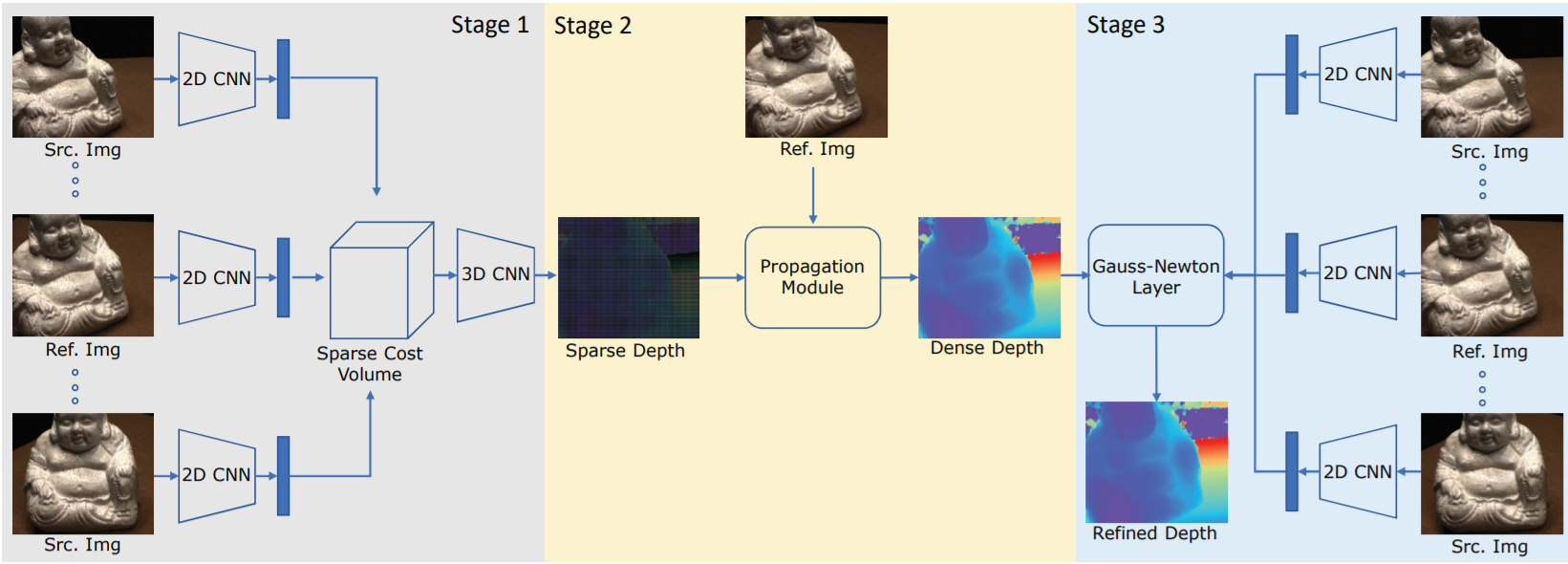}}
    \caption{The architecture of FastMVSNet.
    In the first stage, a sparse cost volume is constructed upon 2D CNN features
    and a sparse high-resolution depth map is estimated using a 3D CNN.
    In the second stage, a the depth map is densified via interpolation with the weights predicted with a CNN.
    In the third stage, the depth map is refined with a differentiable Gauss-Newton layer.}
    \label{fig:fastmvs_arch}
\end{figure*}

FastMVSNet solves the problem of depth map estimation in three stages (\cref{fig:fastmvs_arch}).
The first stage is similar to the whole MVSNet, with two differences.
First, to speed up the computations,
FastMVSNet uses a sparse cost volume half the size of that of the original MVSNet in every spatial dimension.
Second, the shallow refinement CNN is not used.

In the second stage, the initial sparse depth map is densified.
The value in the pixel \(p\) is calculated as the weighted sum over \(k\times k\) window of its neighbours \(N(p)\) by
\begin{equation}
    \tilde{d}(p) = \frac{1}{z_p}\sum_{q\in N(p)}d(q)\cdot \omega_{p,q},
\end{equation}
where \(z_p\) is a normalization constant,
and the weights \(\omega_{p,q}\) are calculated from the reference image with a CNN for each pixel individually.
This allows the network to capture local depth dependencies from the reference RGB image.

In the third stage, the dense depth map \(\tilde{d}\) is refined with Gauss-Newton algorithm,
which optimises the following error function
\begin{equation}
    E(p) = \sum^{N}_{i=1}\norm{F_i\parens{p_i'\parens{\tilde{d}}} - F_0(p)}_2,
\end{equation}
where \(F_i\) and \(F_0\) are the deep representations of the source and reference images,
and \(p_i'\) is the position of pixel \(p\) of reference image reprojected to the source image.
The authors show that this operation is differentiable
which allows them to train the whole method end-to-end with the loss function defined by
\begin{equation}
    L = \sum_{p\in p_\mathrm{valid}} \abs{\tilde{d}(p) - \hat{d}(p)} + \abs{\tilde{d}_r(p) - \hat{d}(p)},
\end{equation}
where \(\tilde{d}_r\) is the final refined depth map,
\(\hat{d}\) is the ground truth depth map,
and \(p_\mathrm{valid}\) is the set of pixels with known ground truth value.

To study the effects of the additional low-quality input depth on this method
we incorporated the depth data directly to the input.
We implemented the approach by concatenating the depth to the RGB input image along the channel dimension.
The rest of the architecture remained unchanged.

We also tried running the input depth through a separate 2D CNN feature extractor,
which seemingly resulted in a slightly better train and val performance,
but we did not have time to analyse the results.

\subsection{CasMVSNet}
\label{sec:meth_cas}
\begin{figure*}[t!]
    \centerline{\includegraphics[scale=0.75]{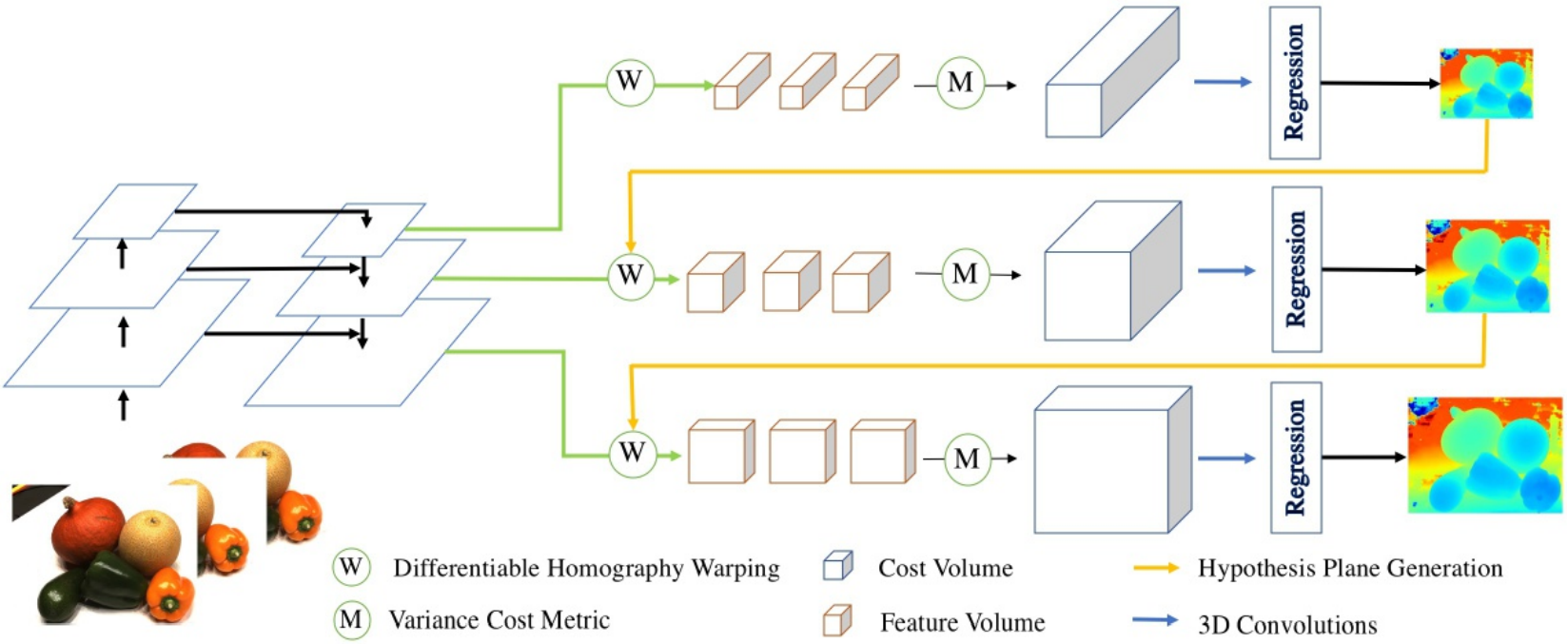}}
    \caption{The architecture of CasMVSNet.
    It can be described as a pyramid of MVSNets,
    where the range of the depth hypotheses used to build the cost volumes at a certain stage
    is based on the depth estimation from the previous stage.}
    \label{fig:casmvs_arch}
\end{figure*}

CasMVSNet can be described as a pyramid of three original MVSNets (\cref{fig:casmvs_arch}),
that compute the cost volumes from different stages in the initial CNN feature extractor,
and produce the depth maps with a gradually increasing resolution.
The key feature of this method is that the range of the depth hypotheses used to build the cost volumes
at stages two and three is based on the depth estimation from the previous stage.

The network is trained end-to-end by the supervision at all stages with the loss function defined by
\begin{equation}
    L = \sum_{k=1}^N \lambda^k \cdot L^k,
\end{equation}
where \(L^k\) is the loss at stage \(k\),
and the weights \(\lambda^k\) equal to \([0.5, 1, 2]\).
The loss at each stage is computed similarly to the original MVSNet as
\begin{equation}
    L^k = \sum_{p\in p_\mathrm{valid}^k} \abs{\tilde{d}^k(p) - \hat{d}^k(p)} + \abs{\tilde{d}^k_r(p) - \hat{d}^k(p)}.
\end{equation}

The stages of CasMVSNet past the first one already use a kind of low quality input depth.
So, to study the effects of additional input depth on this method
we used it as if it was produced by an additional \(0^\mathrm{th}\) stage,
\ie., constraining the range of the depth hypotheses at the first stage based on the input depth.

\section{Experiments}
\label{sec:exper}
\subsection{Data}
\label{sec:exp_data}
\paragraph{DTU.}
Both original methods were trained on DTU dataset~\cite{jensen2014large}.
It was obtained using a camera planted on an industrial robotic arm
and provides accurate camera positions, rotations and distortions.
The dataset captures various installations from a variety of angles
and each in a wide range of lighting conditions from diffuse to direct light from different directions.
The dataset comes with reference point cloud reconstructions of the scenes
obtained with a structured light scanner,
which are used to calculate the ground truth depth maps.

\paragraph{Corrupted DTU.}
Since DTU does not provide the low quality sensor depth,
we used corrupted ground truth data as a simple approximation.
Specifically, we box-downsampled the ground truth depth maps by the factor of 4 and added the noise
with a simplified procedure from~\cite{handa2014benchmark}.
The corruption procedure is summarised by expressions
\begin{equation}
    \begin{gathered}
        d_\mathrm{corrupted} = \frac{b \cdot f}{b \cdot f / d_\mathrm{down} + \mathcal{N}\parens{0, \sigma_d^2} + 0.5},\,
        d_\mathrm{down}(p) = \mathrm{avg}_{q \in N(p)} d(q),
    \end{gathered}
\end{equation}
where \(b = 10 \mathrm{cm}\) is a virtual camera baseline,
\(f = 2892\) is the focal length in pixels for the full resolution \(1600\times 1200\) depth maps in DTU datset,
\(\sigma_d = 1/6\) is the standard deviation of disparity noise as in~\cite{handa2014benchmark},
and \(N(p)\) is the \(4\times 4\) pixel neighbourhood around pixel \(p\).

For both models we used the original train/test/val splits\footnote{
Test set: scans 1, 4, 9, 10, 11, 12, 13, 15, 23, 24, 29, 32, 33, 34, 48, 49,62, 75, 77, 110, 114, 118.
Validation set: scans 3, 5, 17, 21, 28, 35, 37, 38, 40, 43, 56, 59, 66, 67, 82, 86, 106, 117.
Training set: the remaining 79 scans.}
and input resolutions,
\ie., \(640\times 512\) for training and \(1600\times 1200\) for testing.

\subsection{Metrics}
\label{sec:exp_metrics}
To measure the quality of the depth maps during training
we used absolute error of depth map prediction averaged over pixels with nonzero reference depth.

To measure the quality of the final fused point clouds we used standard MVS metrics
\emph{accuracy}, \emph{completeness}, and \emph{overall}.
\emph{Accuracy} is calculated as the mean distance from the reconstruction to the reference and captures the quality of the reconstruction.
\emph{Completeness} is calculated as the mean distance from the reference to the reconstruction,
encapsulating how much of the reference is captured by the reconstruction.
\emph{Overall} is calculated as the mean of the two.
A comprehensive description and discussion of these metrics is given in~\cite[Section~4]{seitz2006comparison}.

\subsection{Implementation}
\label{sec:exp_impl}
For both methods we used the original PyTorch implementations\footnote{
\href{https://github.com/svip-lab/FastMVSNet}{FastMVSNet},
\href{https://github.com/alibaba/cascade-stereo}{CasMVSNet}}.
For generation of the final point clouds we also used the codes from the original repositories of the methods.
For quantitative evaluation of point clouds produced by both methods we utilized the original DTU evaluation toolbox.

Following the FastMVS paper,
we first pretrained the sparse depth map prediction module and propagation module for 4 epochs,
and then trained the whole model end-to-end for 12 epochs.
Finally, we trained the model for other 8 epochs to ensure convergence.
We conducted the experiments using the original values of hyperparameters,
but used 4 GTX1080Ti GPUs instead of 2080Ti.

In addition, our very first attempt lasted for only 8 epochs in total,
and was powered by only one GTX1080Ti GPU,
so the batch size had to be reduced to 4 instead of 16 to fit the memory.
This lead to an unexpected observation, as discussed in \cref{sec:disc}

We trained CasMVSNet with the original hyperparameters,
but used only 4 GTX1080Ti GPUs instead of 8 because of technical reasons,
so our effective batch size was 8 instead of 16.
Since the model incorporates batch normalization layers,
we decided to not use gradient accumulation as a way to increase effective batch size.

\subsection{Results}
\label{sec:exp_results}
\begin{figure*}[t!]
    \centerline{\includegraphics[width=.48 \linewidth]{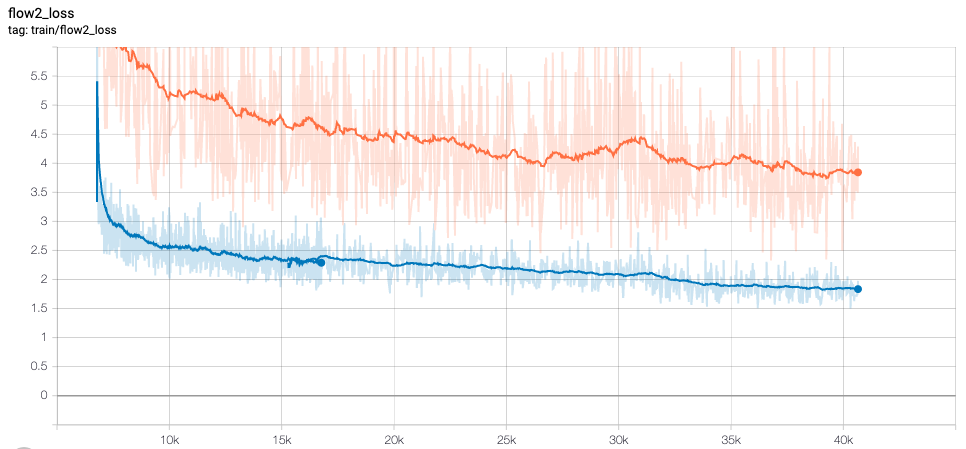}
    \includegraphics[width=.48 \linewidth]{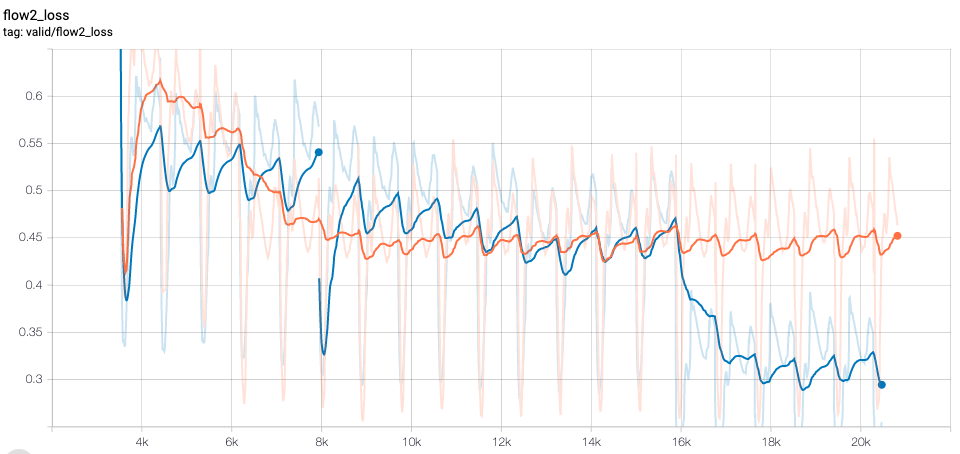}}
    \caption{Train loss on the left and validation absolute depth error on the right for FastMVSNet.
    Our reproduction of the original method is in orange and our modification is in blue.}
    \label{fig:fast_comparison}
\end{figure*}

\begin{figure*}[t!]
    \centerline{\includegraphics[width=.48 \linewidth]{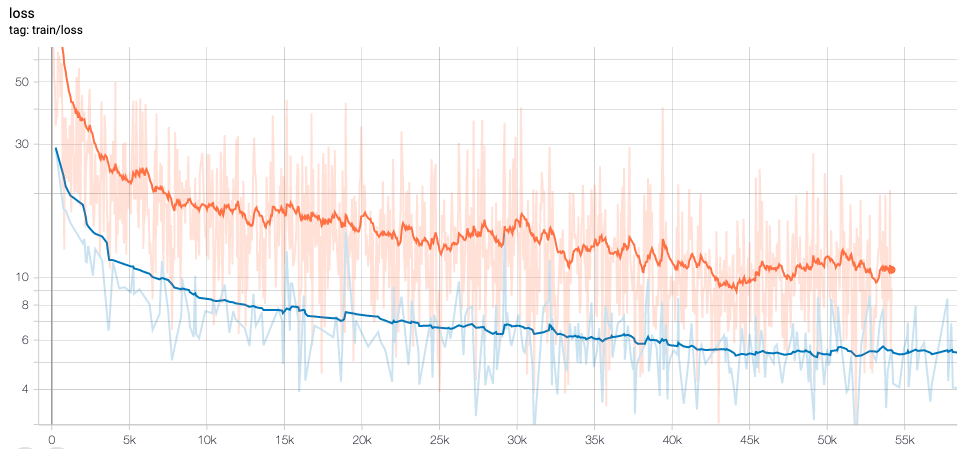}
    \includegraphics[width=.48 \linewidth]{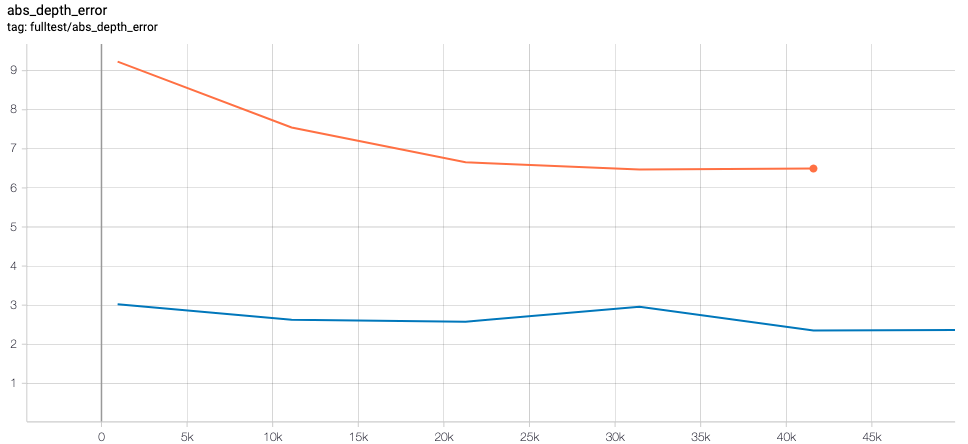}}
    \caption{Train loss on the left and test absolute depth error on the right for CasMVSNet.
    Our reproduction of the original method is in orange and our modification is in blue.}
    \label{fig:cas_comparison}
\end{figure*}

\begin{figure*}[t!]
    \centerline{\begin{tikzpicture}
                    \node[anchor=south west,inner sep=0] (image) at (0,0) {\includegraphics[width=\linewidth]{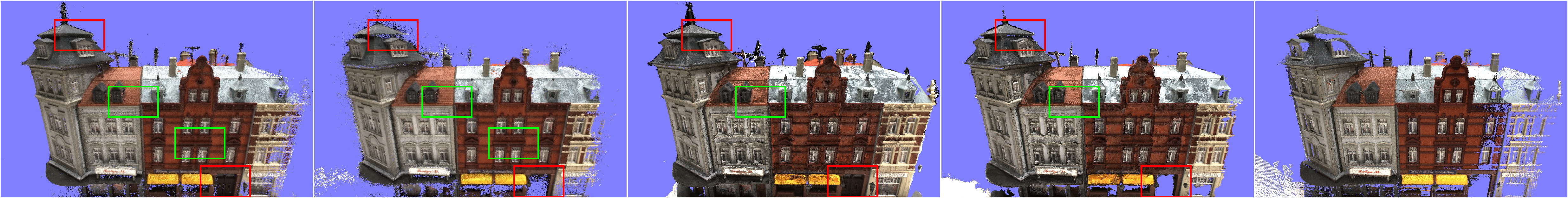}};
                    \node[above=0 of image,inner sep=0] {\includegraphics[width=\linewidth]{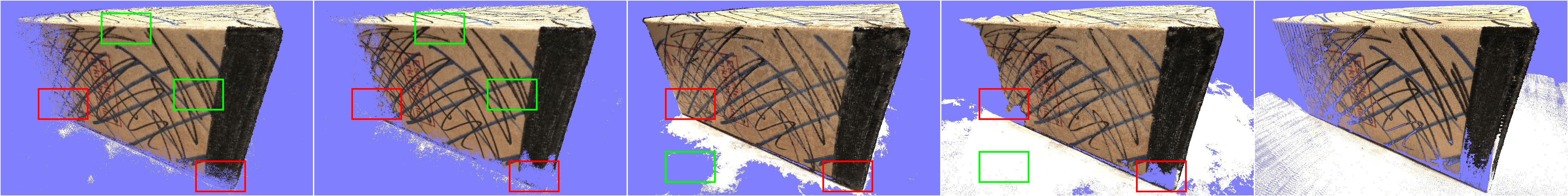}};
                    \begin{scope}[x={(image.south east)},y={(image.north west)}]
                        \footnotesize
                        \def\dx{1/10}
                        \def\y{-2ex}
                        \node at (\dx, \y)    [align=center] {(a) Reproduced FastMVSNet};
                        \node at (3*\dx, \y)  [align=center] {(b) Modified FastMVSNet};
                        \node at (5*\dx, \y)  [align=center] {(c) Reproduced CasMVSNet};
                        \node at (7*\dx, \y)  [align=center] {(d) Modified CasMVSNet};
                        \node at (9*\dx, \y)  [align=center] {(e) Ground truth};
                    \end{scope}
    \end{tikzpicture}}
    \caption{Qualitative results on scans 10 and 9 from DTU.
    The parts of the scene improved with our modification are outlined in green
    and the deteriorated parts are outlined in red.}
    \label{fig:pointclouds}
\end{figure*}
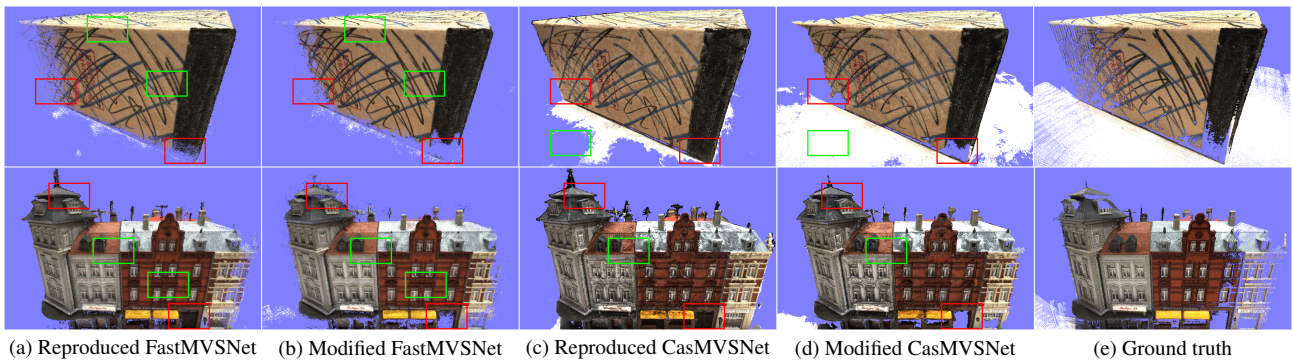

We show the quality measurements for the final point clouds in \cref{tab:dtu_metrics},
the train and validation curves in \cref{fig:fast_comparison,fig:cas_comparison},
and the point clouds generated for two test DTU scans,
scan 10 with simple geometry and scan 9 with more complex geometry,
in \cref{fig:pointclouds}.

\hspace{-2.45em}
\begin{table}[]
    \centerline{
    \setlength\tabcolsep{\widthof{0}*\real{.6}}
    \setlength\aboverulesep{0pt}
    \setlength\belowrulesep{0pt}
    \begin{tabular}{c|c|c|c}
        \toprule
        Method & Accuracy & Completeness & Overall  \\ \hline
        FastMVSNet (orig.) & 0.336 & 0.403 & 0.370 \\
        CasMVSNet (orig.) & 0.346 & 0.351 & 0.348  \\ \hline \midrule
        FastMVSNet (1 GPU) & 0.373 & 0.448 & 0.411 \\
        FastMVSNet (repr.) & 0.417 & 0.467 & 0.442 \\
        CasMVSNet (repr.) & 0.36 & 0.361 & 0.360 \\  \hline \midrule
        FastMVSNet + depth & 0.365 & 0.573 & 0.469 \\
        CasMVSNet + depth & 0.654 & 0.381 & 0.518 \\  \hline \bottomrule
    \end{tabular}
    }
    \caption{The results reported in the original papers,
    the results of the reproduction using the original code,
    and of our experiments with additional depth input.}
    \label{tab:dtu_metrics}
\end{table}

\section{Discussion}
\label{sec:disc}
Both our modifications improve the quality of the output depth maps,
as demonstrated in \cref{fig:fast_comparison,fig:cas_comparison},
but deteriorate the quality of the final fused point clouds,
as shown in \cref{tab:dtu_metrics}.
The modified version of FastMVSNet produces more accurate but less complete point clouds
in comparison to the original one.
The modified version of CasMVSNet produces both less accurate and less complete point clouds
in comparison to the original one.
A possible reason for such mixed results could be the difference in evaluation procedure for depth maps and point clouds.

As demonstrated in \cref{fig:pointclouds},
the reconstructions produced with the modified methods are less complete in comparison to the original methods
in the areas around the holes in the ground truth depth,
and are less noisy in the areas where the ground truth is complete.
This observation may be explained in the following way.

The holes in the ground truth define the holes in the synthesized input depth.
During training, both methods are not penalized for the errors in these holes.
This may result in both modified models learning to rely too much on the input depth,
and producing completely inaccurate depth values if the corresponding pixel in the input depth is missing.
These inaccuracies are filtered out during the fusion into the final point cloud,
which results in a less complete reconstruction.

At the same time, as shown in \cref{fig:pointclouds}~(c, d),
our modified version of CasMVSNet is able to almost fully reconstruct completely textureless white table,
while the original version only reconstructs a small part of it.
We note that all test scenes in DTU possess distinctive textures,
and the table is excluded from evaluation in the original DTU toolbox that we use.

It is worth noting that FastMVSNet that we trained on a single GPU with smaller batch size
and for a reduced number of epochs
counterintuitively yielded point clouds with a higher quality
compared to the one trained on 4 GPUs with the original hyperparameters.

\section{Conclusion}
\label{sec:concl}
In this work we investigated the effects of the additional input low-quality depth
on the results of deep multi-view stereo.
We modified two state-of-the-art deep MVS methods for using with low-quality input depth.
Our experiments show that the additional input depth synthesized via corruption of the ground truth
may improve the quality of deep multi-view stereo.
Investigating the influence of real sensor depth is an important future work direction.

\paragraph{Acknowledgements:}
We acknowledge the usage of Skoltech CDISE HPC cluster Zhores for obtaining the presented results. The work was partially supported by Russian Science Foundation under Grant 19-41-04109.

    \bibliographystyle{spiebib}
    \renewcommand\refname{\uppercase{References}}
    \bibliography{src/bib}
\end{document}